\documentclass[conference]{IEEEtran}
\IEEEoverridecommandlockouts
\usepackage{cite}
\usepackage{amsmath,amssymb,amsfonts}
\usepackage{algorithmic}
\usepackage{graphicx}
\usepackage{textcomp}
\usepackage{xcolor}
\usepackage[latin9]{inputenc}
\usepackage{array}
\usepackage{verbatim}
\usepackage{units}
\usepackage{multirow}
\usepackage{microtype}
\usepackage{booktabs}
\usepackage{amsmath}
\usepackage{amssymb}
\usepackage{graphicx}
\usepackage{bbold}
\usepackage{mathtools}
\def\BibTeX{{\rm B\kern-.05em{\sc i\kern-.025em b}\kern-.08em
    T\kern-.1667em\lower.7ex\hbox{E}\kern-.125emX}}
\begin{document}

\title{ReLaB: Reliable Label Bootstrapping for Semi-Supervised Learning}

\author{\IEEEauthorblockN{Paul Albert, Diego Ortego, Eric Arazo, Noel O'Connor, Kevin McGuinness}
\IEEEauthorblockA{School of Electronic Engineering, \\ 
Insight Centre for Data Analytics, Dublin City Univeristy (DCU)}
\IEEEauthorblockA{paul.albert@insight-centre.org}}

\maketitle

\begin{abstract}
Reducing the amount of labels required to train convolutional neural networks without performance degradation is key to effectively reduce human annotation efforts. We propose Reliable Label Bootstrapping (ReLaB), an unsupervised preprossessing algorithm which improves the performance of semi-supervised algorithms in extremely low supervision settings. Given a dataset with few labeled samples, we first learn meaningful self-supervised, latent features for the data. Second, a label propagation algorithm propagates the known labels on the unsupervised features, effectively labeling the full dataset in an automatic fashion. Third, we select a subset of correctly labeled (reliable) samples using a label noise detection algorithm. Finally, we train a semi-supervised algorithm on the extended subset. We show that the selection of the network architecture and the self-supervised algorithm are important factors to achieve successful label propagation and demonstrate that ReLaB substantially improves semi-supervised learning in scenarios of very limited supervision on CIFAR-10, CIFAR-100 and mini-ImageNet. We reach average error rates of $\boldsymbol{22.34}$ with 1 random labeled sample per class on CIFAR-10 and lower this error to $\boldsymbol{8.46}$ when the labeled sample in each class is highly representative.
Our work is fully reproducible: {https://github.com/PaulAlbert31/ReLaB}.
\end{abstract}
\section{Introduction}
Convolutional neural networks (CNNs) are now the established standard
for visual representation learning~\cite{2018_ECCV_DeepLabv3,2016_CVPR_ResNet,2016_arXiv_Wide},
yet one of their most prevalent limitations is the large quantity of labeled data they require. Although enormous quantities of unlabeled data are now accessible and can be collected with minimal effort, the annotation process remains limited by human intervention~\cite{2018_ECCV_EpicKitchen,2016_arXiv_VisualGenome,2018_ECCV_YouTube-VOS,2018_arXiv_VisDrone}.

There are several alternatives in the literature, that reduce the need for the strong supervision required to train deep neural networks. These include transfer learning~\cite{2018_CVPR_Taskonomy} or few-shot learning~\cite{2018_CVPR_LargeScaleDiff}, where supervised pre-trained features are exploited; semi-supervised learning~\cite{2018_NIPS_Realistic}, where only a part of the dataset is labeled; self-supervised learning~\cite{2019_CVPR_Revis}, where a pretext task is used to learn meaningful features from the data; and label noise~\cite{2017_arXiv_WebVision}, where labels are inferred automatically.

There exists different approaches for semi-supervised scenarios in the state-of-the-art. In particular, consistency regularization~\cite{2019_NIPS_MixMatch,2017_NIPS_MeanTeachers} and pseudo-labeling methods~\cite{2018_ECCV_SSLtransductive,2019_arXiv_Pseudo} are the two dominating strategies. To learn from unlabeled data, consistency regularization encourages consistency in the predictions for the same sample under different perturbations, while pseudo-labeling generates pseudo-labels for unlabeled samples directly from the network predictions.
Despite recent efforts in the semi-supervised learning literature aiming at reducing human supervision further, extreme label scarcity is still  challenging~\cite{2020_ICLR_ReMixMatch,2020_arXiv_FixMatch}. In the absence of labels, the self-supervised paradigm for unsupervised visual representation learning has recently gained traction~\cite{2020_ICLR_SelfLab,2015_ICCV_Context,2019_CVPR_Decoup,2018_ICLR_Rotation, 2020_ICML_SimCLR}. Self-supervised learning constructs a supervisory signal using a pretext task where pretext labels are generated from the data. By solving pretext tasks such as colorization of greyscale images~\cite{2016_ECCV_Colorful}, prediction of image rotations~\cite{2018_ICLR_Rotation}, or contrasting different views of the same image \cite{2020_ICML_SimCLR}, high quality features can be learned without human annotations.
The success of self-supervised learning has motivated its adoption for semi-supervised learning, which improved performance in cases of very low label availability~\cite{2019_arXiv_EnAET,2020_ICLR_ReMixMatch}. Berthelot et al.~\cite{2020_ICLR_ReMixMatch} and Wang et al.~\cite{2019_arXiv_EnAET} use self-supervision as a regularization which stabilizes network training, while Rebuffi et al.~\cite{2019_arXiv_Ziss} make use of self-supervision~\cite{2018_ICLR_Rotation} as an initialization strategy for semi-supervised training. \\
In this paper, we explore the idea of automatically annotating image data using label propagation. In particular, we use representations learned by self-supervised tasks together with a low amount of labels to apply label propagation and spread the available labels to the entirety of the samples. The resulting is a fully labeled dataset which contains numerous incorrect (noisy) annotations. We then select a trusted, clean subset from this noisy dataset that reliably extends the initially labeled data. The extended labeled dataset is then used to enhance the performance of any semi-supervised image classification algorithm when very few labeled samples are available. We name this label bootstrapping strategy ReLaB. When ReLaB is used to bootstrap labels for ReMixMatch \cite{2020_ICLR_ReMixMatch} on CIFAR-10 with 10, 40, 100 labeled samples, we reduce the accuracy error by more than 36, 22, 15 absolute points respectively. ReLaB{'}s unsupervised knowledge-bootstrapping pipeline makes use of self-supervised, image retrieval and label noise solutions to provide an approach for scenarios of extremely scare annotations in semi-supervised learning. This could include visual domains where annotations are either time-consuming and expensive to gather or when expert annotators are required.
Our contributions are as follow:
\begin{enumerate}
    \item We propose an unsupervised knowledge-bootstrapping pipeline which enhances the performance of semi-supervised algorithms when very few labeled samples are available.
    \item We propose a reliable sample selection method in the presence of label noise induced by label propagation. The method is robust to class and noise imbalance.
    \item We evaluate the importance of good self-supervised features for label propagation, and demonstrate the superiority of our approach when dealing with feature-based label noise generated by label propagation.
\end{enumerate}

\section{Related Work}


\subsection{Semi-supervised learning}
Semi-supervised learning seeks to reduce human supervision by jointly learning from sparsely
labeled data and extensive unlabeled data. Semi-supervised learning
has evolved rapidly in recent years by exploiting two distinct 
strategies~\cite{2018_NIPS_Realistic}: consistency regularization and pseudo-labeling.
\paragraph{Consistency regularization} promotes consistency in the network's predictions
for the same unlabeled sample altered by different perturbations.
Notable examples of consistency regularization algorithms are VAT~\cite{2017_TPAMI_VAT}
where samples are perturbed by virtual adversarial attacks, Mean Teacher\cite{2017_NIPS_MeanTeachers}
where a teacher network is built from the exponential moving average of the student network weights to produce perturbed predictions, and ICT~\cite{2019_IJCAI_ICT} which encourages predictions of interpolated samples
to be consistent with the interpolation of the predictions. Berthelot et al. propose MixMatch~\cite{2019_NIPS_MixMatch}, where perturbed predictions are generated by means of data-augmented, sharpened labels and where labeled and unlabeled examples are mixed together using Mixup~\cite{2018_ICLR_Mixup}. MixMatch was extended in ReMixMatch~\cite{2020_ICLR_ReMixMatch} by exploiting distribution alignment~\cite{1992_NIPS_DisAl} and an augmentation anchoring policy.

\paragraph{Pseudo-labeling} directly exploits the network predictions on unlabeled samples by using them as labels (pseudo-labels) to regularize training. Lee et al.~\cite{2013_WREPL_PseudoLabelorri} propose an early attempt at pseudo-labeling, limited to a finetuning stage on a pre-trained network. Shi et al.~\cite{2018_ECCV_MinMax} derive certainty weights for unlabeled samples from their distance to neighboring samples in the feature space. Arazo et al.~\cite{2019_arXiv_Pseudo} have shown that a pure pseudo-labeling without using consistency regularization
can reach competitive performance when addressing confirmation bias~\cite{2019_arXiv_ConfB}. Interestingly, Iscen et al.~\cite{2019_CVPR_LabelProp} proposed a label-propagation based strategy for semi-supervised learning. In particular, they estimate pseudo-labels using both the network prediction and label-propagation on the current features of the network, producing two different supervised objectives.

\subsection{Self-supervised learning}
Self-supervised learning defines proxy or pretext tasks to learn useful representations without human intervention~\cite{2019_CVPR_Revis}. Context prediction~\cite{2015_ICCV_Context}, colorization~\cite{2016_ECCV_Colorful}, puzzle solving \cite{2016_ECCV_Jigsaw}, instance discrimination~\cite{2018_CVPR_NPID}, image rotation prediction~\cite{2018_ICLR_Rotation}, interactive clustering~\cite{2018_ECCV_DeepCluster}, optimal transport~\cite{2020_ICLR_SelfLab}, image transformation  prediction~\cite{2019_CVPR_AET} and construction of local neighborhoods~\cite{2019_ICML_AND} are some examples of pretext tasks. Unsupervised contrastive learning has recently emerged as the new  standard for representation learning~\cite{2020_ICML_SimCLR,2021_ICLR_iMix} where a given sample is encouraged to have similar features to augmented versions of itself and dissimilar representations to other samples in the dataset.

Recent contributions shows that coupling self-supervised and semi-supervised learning can increase the accuracy when few labels are available. Rebuffi et al.~\cite{2019_arXiv_Ziss} use RotNet~\cite{2018_ICLR_Rotation} as a network initialization strategy, ReMixMatch~\cite{2020_ICLR_ReMixMatch} exploits RotNet~\cite{2018_ICLR_Rotation} together with their semi-supervised algorithm to achieve stability with few labels, and EnAET~\cite{2019_arXiv_EnAET} leverage transformation encoding from AET~\cite{2019_CVPR_AET} to improve the consistency of predictions on transformed images.

\subsection{Label propagation for semi-supervised learning}
Label propagation processes stem from the image retrieval literature. Diffusion~\cite{2013_CVPR_DiffRevis,2002_NIPS_LabdiffRandomWalks,2003_NIPS_LGConst} constructs a pairwise affinity matrix, relating images to each other using meaningful features before diffusing the affinity values to the entirety of the graph. In the case of label propagation, the image retrieval objective is reformulated as a label propagation objective which transfers the information from labeled data to an unlabeled dataset~\cite{2006_?_Labelpropcriterion}. The diffusion result can be directly used to estimate labels and finetune pre-trained networks in few-shot learning~\cite{2018_CVPR_LargeScaleDiff} or to define pseudo-labels for semi-supervised learning~\cite{2019_CVPR_LabelProp}. Other attempts using label propagation for semi-supervised learning include dynamically capturing the manifold{'}s structure and regularize it to form compact clusters which facilitate class separation~\cite{2018_ICML_CompactSSL} or to encourage random walks ending in the same class they started from, while penalizing different class endings~\cite{2017_CVPR_Lassociation}.

\subsection{Label noise}
Label noise is a topic of increasing interest for the community~\cite{2020_arXiv_LabelNoiseSurvey}, which aims at limiting the degradation of CNN representations when learning in label noise conditions~\cite{2017_ICLR_Rethinking}. 
Label noise algorithms can be categorized in four different approaches: loss correction~\cite{2018_NIPS_CoTeaching,2017_CVPR_ForwardLoss,2014_arXiv_BootStrap}, relabeling~\cite{2018_CVPR_JointOpt,2019_CVPR_JointOptimizImproved, 2020_arXiv_MetaSoftApple}, semi-supervised~\cite{2018_WACV_SemiSupNoise,2020_ICLR_DivideMix} and regularization~\cite{2018_ICLR_Mixup,2020_NeurIPS_EarlyReg}.
Loss correction algorithms reduce the contribution of the incorrect or noisy labels in the training objective by approximating true labels at sample~\cite{2014_arXiv_BootStrap,2019_ICML_BynamicBootstrapping,2020_NeurIPS_EarlyReg} or class level~\cite{2018_NIPS_GoldLoss, 2017_CVPR_ForwardLoss} or by weighing down noisy samples in the loss~\cite{2018_NIPS_CoTeaching,2018_CVPR_Iterative, 2020_ICML_MentorMix}. Relabeling methods~\cite{2018_CVPR_JointOpt, 2020_arXiv_MetaSoftApple} iteratively update noisy labels to an estimation of the true label.
Semi-supervised methods detect the noisy samples before discarding their labels and exploit the resulting unlabeled content in a semi-supervised setup~\cite{2018_WACV_SemiSupNoise,2019_ICCV_NegativeLearning,2020_ICPR_Robust,2020_ICLR_DivideMix}.
Finally, strong regularization such as Mixup~\cite{2018_ICLR_Mixup} enables robustness to label noise without explicitly addressing it.
A recurrent paradigm used to identify clean samples is the small loss trick~\cite{2019_ICML_BynamicBootstrapping,2018_NIPS_CoTeaching,2020_ICPR_Robust,2018_CVPR_JointOpt} where clean samples exhibit a lower loss early in the training since they are often easier to learn.
 
 \begin{figure*}[!t]
\begin{centering}
\includegraphics[width=2\columnwidth]{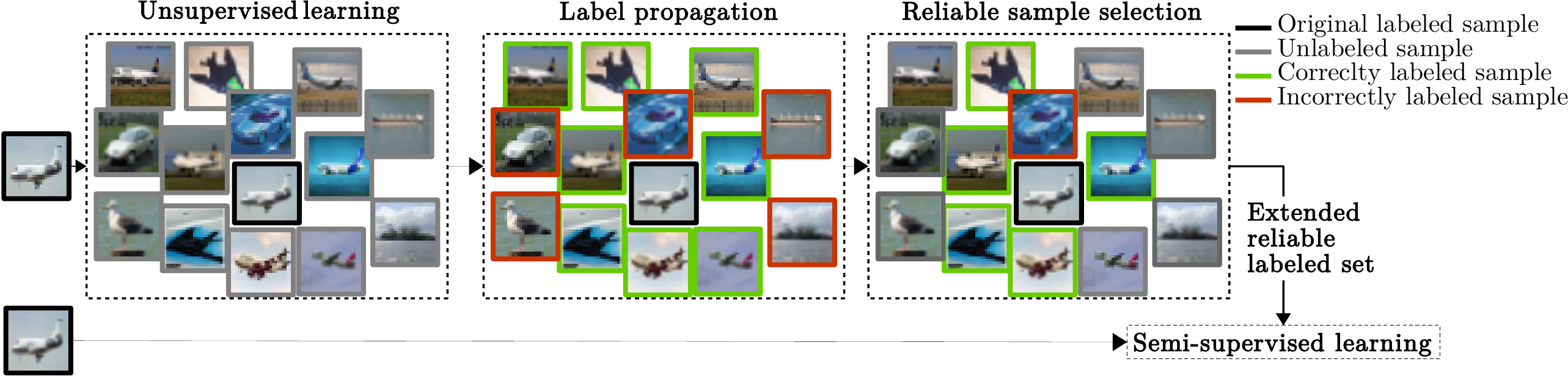}\caption{Reliable Label Bootstrapping (ReLaB) overview (best viewed in color). Unlike traditional SSL (bottom) that directly uses the labeled examples provided (\textit{airplane}), ReLaB (top) bootstraps additional labels before applying SSL. Unsupervised learning using labeled (black) and unlabeled (gray) samples is done to obtain discriminative representations. Label propagation on unsupervised representations propagates the few labeled examples to all samples. This leads to both correct (green) and incorrect (red) labels. A sample selection is finally performed to avoid noisy labels (some will unavoidably be selected) and create a reliable extended labeled set. \label{fig:Paver}}
\par\end{centering}
\end{figure*}

\section{Reliable label bootstrapping for semi-supervised learning\label{sec:algo}}

We formulate a semi-supervised classification task for $C$ classes as learning a model $h_{\psi}$ given a training set $\mathcal{D}$ of $N$ samples. The dataset consists of the labeled set $\mathcal{D}_{l}=\left\{ \left(x_{i},y_{i}\right)\right\} _{i=1}^{N_{l}}$ with corresponding one-hot encoded labels $y_{i}\in\left\{ 0,1\right\} ^{C}$ and the unlabeled set $\mathcal{D}_{u}=\left\{ x_{i}\right\} _{i=1}^{N_{u}}$, $N=N_{l}+N_{u}$ the total number of samples. We consider a CNN for $h_{\psi}:\mathcal{D\rightarrow}\left[0,1\right]^{C}$, where $\psi$ denotes the model{'}s parameters. The network is comprised of a feature extractor $h_{\psi_{f}}:\mathcal{D}\rightarrow\Phi$
with parameters $\psi_{f}$, mapping the input space to the feature
space $\Phi$, and a classifier $h_{\psi_{c}}:\Phi\rightarrow\left[0,1\right]^{C}$
with parameters $\psi_{c}$. \\
We address the case where $\mathcal{D}_l$ contains a low amount of samples. We propose to extend $\mathcal{D}_l$ to a larger dataset $\mathcal{D}_r$ of size $N_r > N_l$ by automatically labeling samples from $\mathcal{D}_u$. To do so, we propagate labels from $\mathcal{D}_l$ to $\mathcal{D}_u$ using self-supervised features learned on $\mathcal{D}$. We build $\mathcal{D}_r$ by selecting clean (reliable) samples from the propagated labels, using label noise methodologies. Training on $\mathcal{D}_r$ greatly improves the performance of semi-supervised algorithms when very few labels are available. Figure~\ref{fig:Paver} presents and overview of our proposed approach.

\subsection{Label propagation on self-supervised features
\label{subsec:Label-propagation}}

Knowledge transfer from the labeled set $\mathcal{D}_{l}$ to the unlabeled set $\mathcal{D}_{u}$ is implicitly done by semi-supervised learning approaches as the network predictions for $\mathcal{D}_{u}$ can be seen as estimated labels. With few labeled samples however, it is difficult to learn useful initial representations from $\mathcal{D}_{l}$ and performance is substantially degraded~\cite{2020_ICLR_ReMixMatch} (see Subsection~\ref{subsec:Very-low-levels}).

Conversely, we propose to learn a set of descriptors in an unsupervised manner and subsequently propagate the labels on the data manifold, in order to retrieve additional labels for the unlabeled data. 
We adopt the established graph diffusion algorithm~\cite{2019_CVPR_LabelProp,2013_CVPR_DiffRevis,2002_NIPS_LabdiffRandomWalks,2017_CVPR_DiffMani,2013_ICCV_DiffAgg} for label propagation. We formulate the label propagation problem in a similar fashion than~\cite{2019_CVPR_LabelProp} except that we study the estimation of $\tilde{y}$ as a label propagation task using unsupervised visual representations learned from all samples in $\mathcal{D}$. In particular, we learn a feature extractor $h_{\varphi_{f}}$ using self-supervision to obtain class-discriminative image representations~\cite{2019_CVPR_Revis} and subsequently propagate labels from the $N_{l}$ labeled images to estimate labels $\tilde{y}$ for the $N_{u}$ unlabeled samples. We do so by solving a label propagation problem based on graph diffusion~\cite{2019_CVPR_LabelProp}. First, the set of descriptors $\left\{ v_{i}\right\} _{i=1}^{N}$ are used to define the affinity matrix:
\begin{align}
S & =D^{-1/2}AD^{-1/2},
\end{align}
where $D=\text{diag}\left(A \mathbb{1}_{N}\right)$ is the degree matrix of the graph and the adjacency matrix $A$ is computed as $A_{ij}=\left(\nicefrac{v_{i}^Tv_{j}}{\left\Vert v_{i}\right\Vert \left\Vert v_{j}\right\Vert  }\right)^{\gamma}$ if $i\ne j$ and $0$ otherwise. ${\gamma}$ weighs the affinity term to control the sensitivity to far neighbors and is set to 3 as in \cite{2019_CVPR_LabelProp}. The diffusion process estimates the $N\times C$ matrix as:
\begin{align}
F & =\left(I-\alpha S\right)^{-1}Y,\label{eq: Diffusion}
\end{align}
where $\alpha$ denotes the probability of jumping to adjacent vertices in the graph and $Y$ is the $N\times C$ label matrix defined such that $Y_{ic}=1$ if sample $x_{i}\in\mathcal{D}_{l}$ and $y_{i}=c$ (i.e. belongs to the $c$ class), where $i$ ($c$) indexes the rows (columns) in $Y.$ Finally, the estimated one-hot label $\tilde{y}_{i}$ is:
\[
\tilde{y}_{ic}=\begin{cases}
1, & \text{{if}}\:c=\underset{c}{\arg\max}\:\:F_{ic}\\
0, & \text{{otherwise}}
\end{cases},
\]
for each unlabeled sample $x_{i}\in\mathcal{D}_{u}$. The estimated labels allow the creation of the extended dataset with estimated noisy labels $\mathcal{\tilde{D}}=\left\{ \left(x_{i},\tilde{y}_{i}\right)\right\} _{i=1}^{N}$, where $\tilde{y}_{i}=y_{i}$, $\forall\:x_{i}\in\mathcal{D}_{l}$.

\subsection{Reliable sample selection: dealing with noisy labels\label{subsec:labelnoise}}

Propagating existing labels using self-supervised representations as described in Section~\ref{subsec:Label-propagation}, results in estimated labels $\tilde{y}_{i}$ that might be incorrect, i.e. label noise. Using noisy labels as a supervised objective on $\mathcal{\tilde{D}}$ leads to performance degradation due to label noise memorization~\cite{2017_ICLR_Rethinking, 2019_CVPR_LabelProp}. Since the label noise in $\mathcal{\tilde{D}}$ comes from features extracted from the data, noisy samples tend to be visually similar to the seed samples which poses a challenging scenario as noise-robust, state-of-the-art training strategies~\cite{2019_ICML_BynamicBootstrapping,2020_NeurIPS_EarlyReg,2018_ICLR_Mixup} experience important limitations (see Table~\ref{tab:directtraining}).
Moreover, we find that this label noise is unbalanced in terms of number of samples and different levels of noise in each class. We report in Table \ref{tab:Imbalance} the median and standard deviation for the number of sample per class (\#samples) and noise ratio over the classes of CIFAR-10 and CIFAR-100 for different amounts of labeled samples in $N_l$. Using the small loss trick to select a subset of clean samples is commonly used in the label noise literature~\cite{2018_WACV_SemiSupNoise,2019_ICCV_NegativeLearning,2020_ICPR_Robust,2020_arXiv_MOIT}, but the issues specific to label noise resulting from label propagation are not addressed in the label noise literature and pose additional challenges, see Section \ref{subsec:Training-directly-on}.

In particular, we identify clean samples using the cross-entropy loss:
\begin{align}
{\ell_{i}}=  -{\tilde{y}_{i}^{T}\log h_{\psi}(x_{i})},\label{eq:weightedCE}
\end{align}
with softmax-normalized logits $h_{\psi}(x_{i})$ and training with a high learning rate (small loss) which helps prevent label noise memorization~\cite{2019_ICML_BynamicBootstrapping} on the extended dataset $\mathcal{\tilde{D}}$.
The reliable set $\mathcal{\mathcal{D}}_{r}=\left\{ \left(x_{i},\tilde{y}_{i}\right)\right\} _{i=1}^{N_{r}},$
with $N_{r}>N_{l}$, is then created by selecting for each class
$c$ the $N_{l}^c$ originally labeled samples for that class $c$
in $\mathcal{\mathcal{D}}_{l}$ and the $N_{r}^{c}-N_{l}^{c}$ samples
in class $c$ from $\mathcal{\mathcal{D}}_{u}$ with the lowest loss
$\ell_{i}$.

Differently from previous works tackling synthetic noise \cite{2020_ICPR_Robust}, we find that the noise present in $\mathcal{\tilde{D}}$ makes the clean sample retrieval using the loss $\ell_{i}$ during any particular epoch unstable and that the noise is class-unbalanced (see Table \ref{tab:Imbalance}), making it more challenging. We therefore impose the selection of a class-balanced clean subset and choose to average the network losses over the last $T$ training epochs. This results in a clean, trusted subset which limits the label noise bias introduced to the semi-supervised algorithm. Table~\ref{tab:Sensitivityspc} shows that the knowledge we bootstrap in $\mathcal{D}_r$ is not overly sensitive to $N_r$.

\begin{table}[t]
\caption{Class and noise imbalance after applying label propagation\label{tab:Imbalance}}
\centering{}{}%
\setlength{\tabcolsep}{3.0pt}
\global\long\def\arraystretch{0.9}%
\resizebox{.45\textwidth}{!}{%
\begin{tabular}{lc>{\centering}c>{\centering}c>{\centering}c>{\centering}c}
\toprule
& \multicolumn{2}{c}{CIFAR-10} & \multicolumn{2}{c}{CIFAR-100} \tabularnewline
{$\frac{N_l}{C}$} & {\#sample} & {noise ratio} & {\#sample} & {noise ratio}\tabularnewline
\cmidrule(lr){1-1}\cmidrule(lr){2-3}\cmidrule(lr){4-5}

{4} & {$4249\pm1726$} & {$24.14\pm10.42$} & {$472\pm161$} & {$50.52\pm16.79$}\tabularnewline

{10} & {$4888\pm1367$} & {$24.28\pm7.43$} & {$477\pm180$} & {$39.92\pm15.31$}\tabularnewline

{25} & {$4990\pm1036$} & {$9.50\pm6.90$} & {$444\pm233$} & {$33.39\pm12.55$}\tabularnewline
\bottomrule
\end{tabular}}{\small\par}
\end{table}


\begin{table*}[t]
\caption{\label{tab:Quality-of-the-1}Label noise percentage in $\mathcal{\tilde{D}}$ using different amounts of labeled samples per class after label propagation using different self-supervised methods and network architectures. Lower is better.}

\centering{}\setlength{\tabcolsep}{0.5pt}
\global\long\def\arraystretch{0.9}%
\resizebox{0.9\textwidth}{!}{{{}}%
\begin{tabular}{p{1.5cm}>{\centering}p{2cm}>{\centering}p{2cm}>{\centering}p{2cm}>{\centering}p{2cm}>{\centering}p{2cm}>{\centering}p{2cm}>{\centering}p{2cm}}
\toprule
 &  & \multicolumn{3}{c}{{CIFAR-10}} & \multicolumn{3}{c}{{CIFAR-100}}\tabularnewline

&  & {1} & {4} & {10} & {4} & {10} & {25}\tabularnewline
\cmidrule(r){0-1}\cmidrule(lr){3-5}\cmidrule(lr){6-8}
\multirow{3}{*}{{RotNet \cite{2018_ICLR_Rotation}}} & {WRN-28-2} & {$67.90\pm8.51$} & {$51.68\pm3.03$} & {$50.09\pm2.55$} & {$83.08\pm0.52$} & {$76.31\pm0.33$} & {$67.81\pm0.15$}\tabularnewline
 & {RN-18} & {$66.02\pm5.98$} & {$53.58\pm1.57$} & {$47.60\pm3.51$} & {$80.83\pm0.56$} & {$73.79\pm0.42$} & {$65.58\pm0.34$}\tabularnewline
 & {RN-50} & {$80.52\pm30.08$} & {$77.58\pm3.45$} & {$71.07\pm1.05$} & {$80.75\pm0.23$} & {$72.33\pm0.15$} & {$62.78\pm0.12$}\tabularnewline
\cmidrule(r){0-1}\cmidrule(lr){3-5}\cmidrule(lr){6-8}
\multirow{3}{*}{{NPID \cite{2018_CVPR_NPID}}} & {WRN-28-2} & {$68.72\pm1.51$} & {$56.3\pm2.42$} & {$51.35\pm1.55$} & {$84.02\pm0.30$} & {$76.91\pm0.40$} & {$67.97\pm0.13$}\tabularnewline
 & {RN-18} & {$59.34\pm7.13$} & {$42.70\pm2.32$} & {$37.14\pm0.48$} & {$77.80\pm0.55$} & {$69.54\pm0.25$} & {$61.29\pm0.67$}\tabularnewline
 & {RN-50} & {$59.44\pm3.10$} & {$44.54\pm2.32$} & {$38.13\pm0.63$} & {$76.67\pm0.58$} & {$68.54\pm0.10$} & {$60.46\pm0.16$}\tabularnewline
\cmidrule(r){0-1}\cmidrule(lr){3-5}\cmidrule(lr){6-8}
\multirow{3}{*}{{UEL \cite{2019_CVPR_UEL}}} & {WRN-28-2} & {$60.81\pm6.41$} & {$45.84\pm2.09$} & {$41.30\pm2.00$} & {$79.21\pm0.09$} & {$71.29\pm0.39$} & {$62.89\pm0.19$}\tabularnewline
 & {RN-18} & {$52.02\pm7.24$} & {$34.51\pm1.03$} & {$29.84\pm0.78$} & {$71.9\pm0.36$} & {$63.25\pm0.41$} & {$56.51\pm0.22$}\tabularnewline
 & {RN-50} & {$49.48\pm7.66$} & {$32.81\pm1.50$} & {$28.78\pm1.08$} & {$69.62\pm0.13$} & {$60.81\pm0.48$} & {$54.08\pm0.22$}\tabularnewline
\cmidrule(r){0-1}\cmidrule(lr){3-5}\cmidrule(lr){6-8}
\multirow{3}{*}{{AND \cite{2019_ICML_AND}}} & {WRN-28-2} & {$61.35\pm0.57$} & {$46.12\pm4.07$} & {$40.78\pm0.27$} & {$79.38\pm0.37$} & {$71.65\pm0.03$} & {$63.29\pm0.38$}\tabularnewline
 & {RN-18} & {$46.55\pm5.64$} & {$28.82\pm1.29$} & {$24.64\pm1.44$} & {$67.48\pm1.04$} & {$58.3\pm0.26$} & {$51.47\pm0.13$}\tabularnewline
 & {RN-50} & $41.96\pm8.74$& $24.34\pm0.94$ & $21.28\pm0.75$ & $66.25\pm0.33$ & $56.6\pm0.52$ & $46.31\pm0.15$\tabularnewline
 \cmidrule(r){0-1}\cmidrule(lr){3-5}\cmidrule(lr){6-8}
 iMix\cite{2021_ICLR_iMix} & {WRN-28-2} & {$53.75\pm2.58$} & {$37.06\pm2.40$} & {$31.27\pm0.27$} & {$76.26\pm0.60$} & {$64.92\pm0.18$} & {$57.95\pm0.45$}\tabularnewline
 + & {RN-18} & {$46.25\pm6.11$} & {${18.55\pm1.81}$} & ${14.51\pm2.35}$& {${49.74\pm1.20}$} & {${42.90\pm0.39}$} & {$39.17\pm0.26$}\tabularnewline
 N-pairs & {RN-50} & $\boldsymbol{38.14\pm8.34}$ & {$\boldsymbol{16.93\pm1.73}$} & $\boldsymbol{13.72\pm1.70}$ & {$\boldsymbol{45.49\pm1.04}$} & {$\boldsymbol{39.41\pm0.08}$} & \textbf{$\boldsymbol{35.75\pm0.26}$}\tabularnewline
\bottomrule
\end{tabular}}
\end{table*}

\subsection{Semi-supervised learning}

Unlike traditional learning from $\mathcal{\mathcal{D}}_{l}$ and $\mathcal{\mathcal{D}}_{u}$, ReLaB provides semi-supervised algorithms with a (larger) reliable labeled set $\mathcal{\mathcal{D}}_{r}$ extended from the original (smaller) labeled set $\mathcal{\mathcal{D}}_{l}$. The extension from $\mathcal{\mathcal{D}}_{l}$ to $\mathcal{\mathcal{D}}_{r}$ is done in a completely unsupervised manner and promotes a significant reduction of the error rates of SSL algorithms when few labels are given, e.g. in Table~\ref{tab:Interpretable-results-1spc} the $50.62$\% error of ReMixMatch~\cite{2020_ICLR_ReMixMatch} in CIFAR-10 for one labeled sample per class ($N_{l}=10$) is reduced to $8.46\%$.

\section{Experiments \label{sec:exp}}

\subsection{Datasets and implementation details\label{subsec:Data_Implementation}}

We experiment with three image classification datasets: CIFAR-10~\cite{2009_CIFAR},
CIFAR-100~\cite{2009_CIFAR}, and mini-ImageNet~\cite{2016_NIPS_MiniImageNet}.
CIFAR (mini-ImageNet) data consists of 60K $32\times32$ ($84\times84$)
RGB images split into 50K training samples and 10K for testing. CIFAR-10
samples are organized in 10 classes, while CIFAR-100 and mini-ImageNet
are in 100.
We follow common practices for image retrieval~\cite{2015_ECCV_Retrieval,2018_TPAMI_ImageRetrieval} and perform PCA whitening as well as $L_{2}$ normalization on the features $v$ before applying diffusion.
We construct the reliable set $\mathcal{D}_{r}$
by training for 60 epochs with a high learning rate (0.1) to prevent label
noise memorization \cite{2019_ICML_BynamicBootstrapping} and select
the samples with the lowest loss per class at the end of the training. We average the per-sample loss over the last
$T=30$ epochs of training. 
For the semi-supervised learning experiments, we always use a standard WideResNet-28-2~\cite{2016_arXiv_Wide}
for fair comparison with related work. We combine our approach with
state-of-the-art pseudo-labeling~\cite{2019_arXiv_Pseudo} and consistency
regularization-based~\cite{2020_ICLR_ReMixMatch} semi-supervised methods to prove the stability of ReLaB when applied to different semi-supervised strategies. We use the default configuration for pseudo-labeling\footnote{https://github.com/EricArazo/PseudoLabeling} except for the network initialization, where we make use of the Rotation self-supervised objective~\cite{2018_ICLR_Rotation} and freeze all the layers up to the last convolutional block in a similar fashion to Rebufi et al.~\cite{2019_arXiv_Ziss}. We find that this is necessary to preserve strong early features throughout the training.
The network is warmed up on the labeled set for $200$ epochs and then trained for $400$ epochs on the whole dataset. For ReMixMatch\footnote{https://github.com/google-research/remixmatch} we train the network from scratch for $256$ epochs.
Experiments in Section~\ref{subsec:Training-directly-on} for the supervised alternatives on dealing with label noise~\cite{2019_ICML_BynamicBootstrapping,2018_ICLR_Mixup} follow the authors's configurations, while cross-entropy and Mixup training in Table \ref{tab:directtraining} is done for 150 epochs with an initial learning rate of 0.1 that we divide by 10 in epochs 80 and 130.
\subsection{Self-supervised representations for label propagation\label{subsec:expLabprop}}
Label propagation relies upon self-supervised representations extracted form the data, i.e.~the quality of the propagation directly depends on these representations. We propose to explore different unsupervised learning alternatives to obtain these representations. Table~\ref{tab:Quality-of-the-1}, presents the label noise percentage of the extended labeled set $\mathcal{\tilde{D}}$ in CIFAR-10 (100) formed after label propagation of the specified self-supervised representations with 1, 4 and 10 (4, 10 and 25) labeled samples per-class in $\mathcal{D}_{l}$. We select RotNet~\cite{2018_ICLR_Rotation}, NPID~\cite{2018_CVPR_NPID}, UEL~\cite{2019_CVPR_UEL}, AND~\cite{2019_ICML_AND} and iMix~\cite{2021_ICLR_iMix} as five recent self-supervised methods. We experiment training  WideResNet-28-2 (WRN-28-2)~\cite{2016_arXiv_Wide}, ResNet-18 (RN-18) and ResNet-50 (RN-50)~\cite{2016_CVPR_ResNet} architectures. All the self-supervised methods are trained using the recommended configuration. We report average noise percentage and standard deviation for 3 different labeled subset $\mathcal{D}_l$. We confirm that the architecture has a key impact on the label noise percentage, which agrees with previous observations on the quality benefits of self-supervised features from larger architectures~\cite{2019_CVPR_Revis, 2021_ICLR_iMix}. We find that using diffusion on features learned using the iMix algorithm promotes the lowest amount of noise and adopt it together with a ResNet-50 in the subsequent experiments.

\subsection{Dealing with noisy labels \label{subsec:Training-directly-on}}
We analyze the importance of the selected number of samples $N_{r}$ over the label noise percentage in the extended reliable subset  $\mathcal{D}_r$ and semi-supervised performance (using ReMixMatch (RMM)~\cite{2020_ICLR_ReMixMatch}). Table~\ref{tab:Sensitivityspc} shows how, a balance has to be found between a sufficient amount of bootstrapped samples and a low noise ratio. Increasing the number of samples in $\mathcal{D}_r$ is beneficial up to $100$ samples per class, where adding more does not compensate the higher noise percentage. Based on this experiment and the typical amounts of labeled samples needed to perform successful
SSL~\cite{2019_arXiv_Pseudo,2019_NIPS_MixMatch,2019_CVPR_LabelProp,2017_NIPS_MeanTeachers}, we choose a conservative $N_{r}=500\,\left(4000\right)$ for CIFAR-10 (100) for further experiments. 

\begin{table}[]
\caption{\label{tab:Sensitivityspc}Sensitivity of semi-supervised methods to different amounts of bootstrapped samples per class ($\frac{N_r}{c}$) considering an initial 4 labeled samples per class ($\frac{N_{l}}{c}=4$). We report label noise percentage in $\mathcal{D}_{r}$ and final error rates after semi-supervised training.}
\centering{}
\global\long\def\arraystretch{0.9}%
\resizebox{0.48\textwidth}{!}{{{}}%
\begin{tabular}{l>{\centering}p{1.5cm}>{\centering}p{1.5cm}>{\centering}p{1.5cm}>{\centering}p{1.5cm}}
\toprule
 & \multicolumn{2}{c}{{CIFAR-10}} & \multicolumn{2}{c}{{CIFAR-100}}\tabularnewline
 $\frac{N_r}{C}$ & {Noise (\%)} & {SSL error} & {Noise (\%)} & {SSL error}\tabularnewline
\cmidrule(lr){0-0}\cmidrule(lr){2-3}\cmidrule(lr){4-5}
{$25$} & {$\boldsymbol{0.40}$} & {$12.12$} & {$\boldsymbol{25.48}$} & {$51.90$}\tabularnewline
{$50$} & {$0.60$} & {$9.18$} & {$30.20$} & {$51.43$} \tabularnewline
{$75$} & {$1.07$} & {$\boldsymbol{8.76}$} & {$33.51$} & {$\boldsymbol{50.65}$}\tabularnewline
{$100$} & {$1.30$} & {$8.79$} & {$35.69$} & {$51.14$}\tabularnewline
\bottomrule
\end{tabular}}
\end{table}

Since $\mathcal{\tilde{D}}$ is corrupted with label noise, it is reasonable to expect that supervised alternatives on dealing with label noise \cite{2019_ICML_BynamicBootstrapping,2018_ICLR_Mixup} could help combat this label noise. Table~\ref{tab:directtraining}  compares our proposed approach against training on $\tilde{\mathcal{D}}$ with standard cross-entropy (CE) and label noise robust methods such as Mixup (M)~\cite{2018_ICLR_Mixup}, the Dynamic Bootstrapping (DB) loss correction method~\cite{2019_ICML_BynamicBootstrapping} and the Early Regularization (ELR) strategy \cite{2020_NeurIPS_EarlyReg}. We also report using the retrieval score (Ret. score) from the label propagation ($\underset{c}{\max}\:\:F_{ic}$ in eq. \ref{eq: Diffusion}) instead of ReLaB for selecting the trusted subset. In both CIFAR-10 and CIFAR-100, ReLaB + ReMixMatch (RMM) outperforms supervised alternatives. 

\begin{table}[t]
\caption{\label{tab:directtraining} Learning from $\mathcal{\hat{D}}$ constructed from 4 labeled samples per class on CIFAR-10 ($N_{l}=40$) and CIFAR-100 ($N_{l}=400$). Error rates}
\centering{}
\global\long\def\arraystretch{0.9}%
\resizebox{0.45\textwidth}{!}{%
\begin{tabular}{l>{\centering}p{2.0cm}>{\centering}p{2.0cm}}
\toprule
 & {CIFAR-10} & {CIFAR-100}\tabularnewline
\cmidrule(lr){0-0}\cmidrule(lr){2-3}
{CE} & $22.64$ & $59.88$\tabularnewline
{M \cite{2018_ICLR_Mixup}} & $21.27$ & $57.92$\tabularnewline
{DB \cite{2019_ICML_BynamicBootstrapping}} & $14.84$ & $55.07$\tabularnewline
{ELR \cite{2020_NeurIPS_EarlyReg}} & $17.39$ & $47.95$ \tabularnewline
\cmidrule(lr){0-0}\cmidrule(lr){2-3}
{Ret. score + PL \cite{2019_arXiv_Pseudo}} & $17.55$ & $54.19$ \tabularnewline
{ReLaB + PL \cite{2019_arXiv_Pseudo}} & $12.38$ & $53.58$\tabularnewline
{ReLaB + RMM \cite{2020_ICLR_ReMixMatch}} & $\boldsymbol{6.68}$ & $\boldsymbol{43.53}$\tabularnewline
\bottomrule
\end{tabular}}
\end{table}

\begin{table*}[]
\caption{\label{tab:Final-results-of}ReLaB for semi-supervised learning on CIFAR-10 and CIFAR-100 with very limited amounts of labeled data. Error rates. We mark with {$\dagger$} the methods we run ourselves. Other results are from~\cite{2020_arXiv_FixMatch} or~\cite{2019_arXiv_EnAET}. Bold denotes best.}

\centering{}
\global\long\def\arraystretch{0.50}%
\resizebox{1\textwidth}{!}{%
\begin{tabular}{lcccccccc}
\toprule
&\multicolumn{4}{c}{{CIFAR-10}} & \multicolumn{4}{c}{{CIFAR-100}}\tabularnewline
{Labeled samples} & {10} & {40} & {100} & {250} & {100} & {400} & {1000} & {2500}\tabularnewline
\cmidrule(lr){0-0}\cmidrule(lr){2-5}\cmidrule(lr){6-9}
{$\pi$-model \cite{2015_NIPS_PIModel}} & {-} & {-} & {-} & {$54.26\pm3.97$} & {-} & {-} & {-} & {$57.25\pm0.48$} \tabularnewline
{MT \cite{2017_NIPS_MeanTeachers}} & {-} & {-} & {-} & {$32.32\pm2.30$} & {-} & {-} & {-} & {$53.91\pm0.57$} \tabularnewline
{PL \cite{2019_arXiv_Pseudo}$\dagger$} & {$55.61\pm5.28$} & {$29.65\pm5.71$} & {$12.83\pm0.68$} & {$12.00\pm0.32$} &  {$88.23\pm0.32$} & {$67.57\pm0.58$} & {$55.20\pm0.69$} & {$45.42\pm0.68$} \tabularnewline
{MM \cite{2019_NIPS_MixMatch}} & {-} & {$47.54\pm11.50$} & {-} & {$11.05\pm0.86$} & {-} & {$67.61\pm1.32$} & {-} & {$39.94\pm0.37$} \tabularnewline
{UDA \cite{2019_arXiv_UDA}} & {-} & {$29.05\pm5.93$} & {-} & {$8.82\pm1.08$} & {-} & {-} & {-} & {-} \tabularnewline
{RMM \cite{2020_ICLR_ReMixMatch}$\dagger$} & {$58.80\pm1.98$} & {$31.36\pm4.37$} & {$22.56\pm2.58$} & {$7.80\pm0.83$} & {$81.18\pm2.36$} & {$57.44\pm2.53$} & {$44.11\pm1.51$} & {$36.66\pm0.33$}\tabularnewline
{EnAET \cite{2019_arXiv_EnAET}} & {-} & {-} & {$9.35$} & {$7.60\pm0.34$} & {-} & {-} & {$58.73$} & {-}\tabularnewline
\cmidrule(lr){0-0}\cmidrule(lr){2-5}\cmidrule(lr){6-9}
{ReLaB + PL$\dagger$} & {$29.89\pm3.64$} & {$12.38\pm0.78$} & {$11.38\pm0.64$} & {$10.68\pm0.66$} & {$68.04\pm2.52$} & {$53.58\pm1.20$} & {$48.79\pm0.82$} & {$43.84\pm0.72$} \tabularnewline
{ReLaB + RMM$\dagger$} & {$\boldsymbol{22.34\pm4.92}$} & {$\boldsymbol{8.23\pm1.38}$} & {$\boldsymbol{6.89\pm0.18}$} & {$\boldsymbol{6.71\pm0.20}$} & {$\boldsymbol{62.02\pm2.77}$} & {$\boldsymbol{44.09\pm0.51}$} & {$\boldsymbol{39.58\pm0.70}$} & {$\boldsymbol{35.19\pm0.74}$}\tabularnewline
\bottomrule
\end{tabular}}
\end{table*}

\begin{table*}[]
\centering{}\caption{Effect of ReLaB on mini-ImageNet with very limited amounts of labeled data and $N_{r}=4000$. Error rates. \label{tab:Generalization-of-the}}

\global\long\def\arraystretch{0.9}%
\resizebox{0.6\textwidth}{!}{{{}}%
\begin{tabular}{l>{\centering}c>{\centering}c>{\centering}c>{\centering}c}
\toprule
{Labeled samples} & {100} & {400} & {1000} & {2500}\tabularnewline
\cmidrule(lr){0-0}\cmidrule(lr){2-5}
{PL \cite{2019_arXiv_Pseudo}} & \textit{$90.89\pm0.62$} & \textit{$85.00\pm0.94$} & \textit{$75.47\pm0.52$} & \textit{$55.10\pm1.52$}  \tabularnewline
{ReLaB + PL} & $\boldsymbol{76.25\pm0.80}$ & \textit{$\boldsymbol{66.66\pm0.54}$} & \textit{$\boldsymbol{60.82\pm1.04}$} & \textit{$\boldsymbol{52.39\pm1.03}$} \tabularnewline
\bottomrule
\end{tabular}}
\end{table*}

\begin{figure}[t]
\centering{}\includegraphics[width=1\columnwidth]{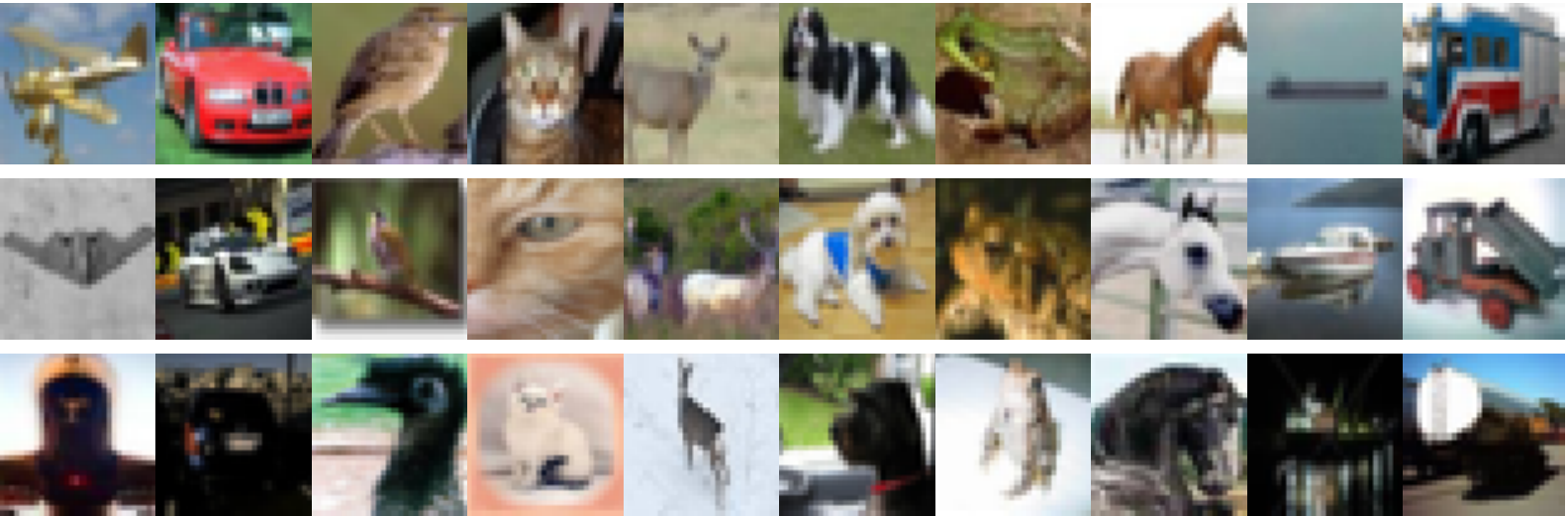}
\caption{Labeled samples used for the 1 sample per class study on CIFAR-10 and taken from~\cite{2020_arXiv_FixMatch}, ordered from top to bottom from most representative to least representative. \label{fig:1spc}}
\end{figure}

\begin{table}[t]
\begin{centering}
\caption{Error rates for 1 sample per class on CIFAR-10 with different labeled sets. We run all the methods ourselves except for FixMatch \cite{2020_arXiv_FixMatch}. Key: MR (Most Representative), LR (Less Representative), NR (Not Representative). \label{tab:Interpretable-results-1spc}}
\centering{}
\global\long\def\arraystretch{0.9}%
\resizebox{0.40\textwidth}{!}{{{}}%
\begin{tabular}{l>{\centering}p{1cm}>{\centering}p{1cm}>{\centering}p{1cm}}
\toprule
 & {MR} & {LR} & {NR}\tabularnewline
\cmidrule(lr){0-0}\cmidrule(lr){2-4}
{ReMixMatch~\cite{2020_ICLR_ReMixMatch}} & {$50.62$} & \textit{$62.57$} & \textit{$90.00$}\tabularnewline
{FixMatch~\cite{2020_arXiv_FixMatch}} & {$22.00$} & {$35.00$} & {$90.00$}\tabularnewline
\cmidrule(lr){0-0}\cmidrule(lr){2-4}
{ReLaB + PL} & \textit{$19.86$} & \textit{$32.38$} & \textit{$79.9$}\tabularnewline
{ReLaB + RMM} & \textit{$\boldsymbol{8.46}$} & \textit{$\boldsymbol{21.75}$} & {$\boldsymbol{78.25}$}\tabularnewline
\bottomrule
\end{tabular}}
\par\end{centering}
\end{table}

\subsection{Semi-supervised learning with ReLaB\label{subsec:Paver_SSL}}
Table~\ref{tab:Final-results-of} presents the benefits of ReLaB
for semi-supervised learning, showing great improvements for both PL~\cite{2019_arXiv_Pseudo} and
ReMixMatch (RMM)~\cite{2020_ICLR_ReMixMatch} when paired with ReLaB. Our focus is on very low levels of labeled samples as semi-supervised methods~\cite{2020_ICLR_ReMixMatch} already achieve very good performance with larger numbers of labeled samples. We further study the 1 sample per class scenario in Section~\ref{subsec:Very-low-levels}.

Table~\ref{tab:Generalization-of-the} demonstrates the scalability of our approach to higher resolution images by evaluating ReLaB + PL~\cite{2019_arXiv_Pseudo} on mini-ImageNet~\cite{2016_NIPS_MiniImageNet}. Due to GPU memory constrains, we use ResNet-18 instead of ResNet-50 to train iMix with an acceptable batch size for the mini-ImageNet experiments.

\subsection{Very low levels of labeled samples\label{subsec:Very-low-levels}}

The high standard deviation using 1 sample per class ($N_{l}=10$) in CIFAR-10 (Table~\ref{tab:Final-results-of}) motivates the proposal of a more reasonable method to compare against other approaches. To this end, Sohn et al.~\cite{2020_arXiv_FixMatch} proposed 8 different labeled subsets for 1 sample per class in CIFAR-10, ordered from more representative to less representative, we reduce the experiments to 3 subsets: the most representative, the least representative, and one in the middle. Figure~\ref{fig:1spc} shows the selected subsets; the exact sample ids are available together with our code for easy reproduction.

Table~\ref{tab:Interpretable-results-1spc} reports the performance for each subset and compares against  FixMatch~\cite{2020_arXiv_FixMatch} and ReMixMatch~\cite{2020_ICLR_ReMixMatch}.
Note that the results obtained for the less representative samples reflect the results that can be expected on average when drawing labeled samples randomly. In the case of the not representative subset, ReLaB enables the semi-supervised learning algorithms to converge better than a random guess. We find that for CIFAR-100 and mini-ImageNet, runs accross different initial labeled samples are more consistent and a comparison to other methods can be made even when drawing the labeled samples at random.

\section{Conclusion}
ReLaB leverages methods from different vision tasks (image retrieval, self-supervised feature learning, label noise for image classification) to propose an unsupervised bootstrapping of additional labeled samples which can in term be used to enhance any semi-supervised learning algorithm. We demonstrate the direct impact of better unsupervised features for the performance of ReLaB and the relevance of our reliable sample selection. Using the extended amount of supervision of ReLaB{'}s reliable set, we enable semi-supervised algorithms to reach remarkable and stable accuracies with very few labeled samples on standard datasets. The extremely low levels of labeled samples we consider in this paper ($<25$ per class) addresses a gap in the semi-supervised literature, which otherwise perform on par with supervised learning for moderate levels of labeled samples ($>25$ per class). Direct applications of ReLaB would include scenarios where the annotation of images is very time consuming or requiring expert annotators for example for medical imaging.

\section*{Acknowledgements}
This publication has emanated from research conducted with the financial support of Science Foundation Ireland (SFI) under grant number [SFI/15/SIRG/3283] and [SFI/12/RC/2289\_P2] as well as from the Department of Agriculture, Food and Marine on behalf of the Government of Ireland under Grant Number [16/RC/3835].

\bibliographystyle{IEEEtran}
\bibliography{egbib}

\end{document}


\title{Reliable Label Bootstrapping for Semi-Supervised Learning - Supplementary material}

\maketitle
\section{Ablation study for label propagation}

We analyze the importance on the final semi-supervised performance of using diffusion and good unsupervised features for label propagation. We compare in Table \ref{tab:Benefits-of-diffusion} the performance achieved when using nearest neighbours compared to diffusion and two different unsupervised pre-training strategies (AND \cite{2019_ICML_AND} and RotNet \cite{2018_ICLR_Rotation}). The results show the importance of estimating good features and proper label propagation strategy.

\begin{table}[b]
\caption{Benefits of diffusion over a nearest neighbor (NN) for label propagation. We report error rates over the same labeled subset and use ReLaB + Pseudo-labeling. \label{tab:Benefits-of-diffusion}}
\centering{}{}%
\global\long\def\arraystretch{0.9}%
\resizebox{0.45\textwidth}{!}{{{}}%
{}%
\begin{tabular}{lc>{\centering}c>{\centering}c}
\toprule
 {Dataset} & {CIFAR-10} & {CIFAR-100}\tabularnewline
\# labeled samples & 40 & 400\tabularnewline
\midrule
{RotNet \cite{2018_ICLR_Rotation} + NN} & $25.64$ & $69.76$\tabularnewline
{AND \cite{2019_ICML_AND} + NN} & $40.70$ & $69.67$\tabularnewline
{RotNet \cite{2018_ICLR_Rotation} + Diffusion} & $16.49$ & $65.35$\tabularnewline
{AND \cite{2019_ICML_AND} + Diffusion} & {$\boldsymbol{14.21}$} & {$\boldsymbol{57.09}$}\tabularnewline
\bottomrule
\end{tabular}}
\end{table}

\section{ReMixMatch training \label{subsec:shorter-training-RMM} }

We report in Table \ref{tab:Ilonger-training} the error rates for ReMixMatch, with and without our proposed ReLaB method, when training for 256 epochs instead of the original 1024 \cite{2020_ICLR_ReMixMatch}. Although longer training is always beneficial, we observe convergence to a reasonable performance in 256 epochs. In the paper we adopt the 256 configuration as it substantially reduces training time.

\begin{table}

\caption{Error rate of short and long training of the ReMixMatch algorithm \cite{2020_ICLR_ReMixMatch}. We report mean and standard deviation over 3 runs.\label{tab:Ilonger-training}}

\centering{}{}%
\setlength{\tabcolsep}{3.0pt}
\global\long\def\arraystretch{0.9}%
\resizebox{0.48\textwidth}{!}{
\begin{tabular}{lc>{\centering}c>{\centering}c>{\centering}c}
\toprule
{Dataset} & {CIFAR-10} & {CIFAR-10} & {CIFAR-100}\tabularnewline
\# labeled samples & 250 & 40 & 400\tabularnewline
\midrule
{ReLaB} & {No} & {Yes} & {Yes}\tabularnewline
{ReMixMatch - 256} & {$7.8\pm0.83$} & {$10.04\pm4.58$} & {$48.59\pm0.7$}\tabularnewline
{ReMixMatch - 1024} & {$6.24\pm0.34$} & {$9.04\pm4.37$} & {$47.24\pm0.68$}\tabularnewline
\bottomrule
\end{tabular}}
\end{table}

\section{Ablation study for Pseudo-labeling}

We study the effect on the pseudo-labeling algorithm in \cite{2019_arXiv_Pseudo} when using an unsupervised initialization with RotNet \cite{2018_ICLR_Rotation} and freezing all the layers up to the last convolutional block to avoid fitting label noise of the reliable extended set $\mathcal{D}_{r}$. Unsupervised initialization and early layers freezing is also adopted in \cite{2019_arXiv_Ziss} to improve pseudo-labeling. We show in Table \ref{tab:AblationstudyPL} that both strategies contribute to better pseudo-labeling performance.

\begin{table}
\caption{Ablation study on the error rate of pseudo-labeling (PL) algorithm in \cite{2019_arXiv_Pseudo} combined with ReLaB when using unsupervised initialization and layer  freezing (LF). \label{tab:AblationstudyPL}}

\centering{}{}%
\setlength{\tabcolsep}{3.0pt}
\global\long\def\arraystretch{0.9}%
\resizebox{.45\textwidth}{!}{%
\begin{tabular}{lc>{\centering}c>{\centering}c}
\toprule
{Dataset} & {CIFAR-10} & {CIFAR-100}\tabularnewline
\# labeled samples & 40 & 400 \tabularnewline\midrule
{ReLaB + PL \cite{2019_arXiv_Pseudo}} & {$22.12$} & {$58.17$}\tabularnewline
\midrule
{ReLaB + PL (RotNet \cite{2018_ICLR_Rotation})
} & {$15.72$} & {$59.04$}\tabularnewline
\midrule
{ReLaB + PL (RotNet  \cite{2018_ICLR_Rotation} + LF)} & {$\boldsymbol{14.21}$} & {$\boldsymbol{57.09}$}\tabularnewline
\bottomrule
\end{tabular}}{\small\par}
\end{table}

\section{Visualization of the bootstrapped samples}
\begin{figure*}[t]
\centering{}\includegraphics[width=1\textwidth]{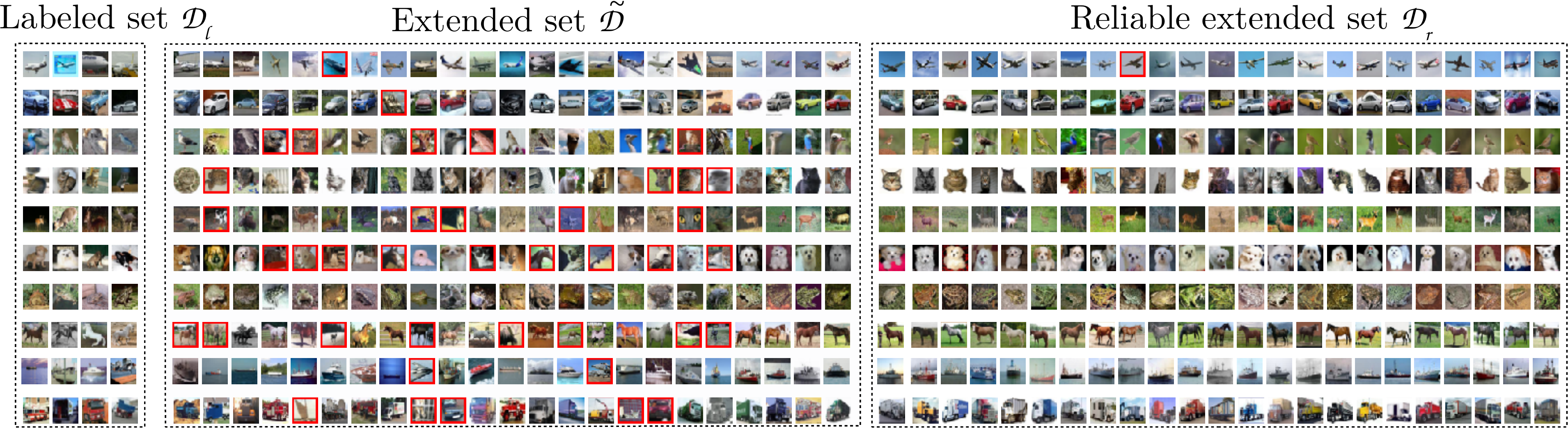}\caption{\label{fig:bootstrap-samples}Qualitative example of label propagation
and reliable sample selection in CIFAR-10 with four seed samples per class. Best viewed on a computer.}
\end{figure*}

Figure \ref{fig:bootstrap-samples} displays the capacity of our reliable sample selection to select an extended clean subset for the semi-supervised algorithm. The first row displays the seed samples; the middle row display a random subset of the samples labeled using label propagation on self-supervised features; the last row displays the reliable samples we select to extend the label set. Images with a red border have a noisy label. The label noise is reduced in the reliable extended set. The figure is best viewed on a computer.

\bibliographystyle{IEEEtran}
\bibliography{egbib}